\documentclass[sigconf]{aamas}  %

\usepackage{booktabs}
\usepackage{graphicx}
\usepackage{amsmath}
\usepackage{xspace}
\usepackage{amssymb}
\usepackage{bm}
\usepackage{color}
\usepackage{url}
\usepackage{booktabs}
\usepackage{subfig}
\usepackage{algorithmic}
\usepackage{natbib}
\usepackage{tikz}
\usetikzlibrary{positioning}
\usepackage{amsthm}

\renewcommand\footnotetextcopyrightpermission[1]{} %
\newcommand\blfootnote[1]{%
  \begingroup
  \renewcommand\thefootnote{}\footnote{#1}%
  \addtocounter{footnote}{-1}%
  \endgroup
}

\setcopyright{ifaamas}  %
\acmDOI{doi}  %
\acmISBN{}  %
\acmConference[AAMAS'19]{Proc.\@ of the 18th International Conference on Autonomous Agents and Multiagent Systems (AAMAS 2019), N.~Agmon, M.~E.~Taylor, E.~Elkind, M.~Veloso (eds.)}{May 2019}{Montreal, Canada}  %
\acmYear{2019}  %
\copyrightyear{2019}  %
\acmPrice{}  %

\renewcommand\footnotetextcopyrightpermission[1]{} %

\begin{document}

\title{Safer Deep RL with Shallow MCTS:\\
A Case Study in Pommerman}

\author{Bilal Kartal$^*$, Pablo Hernandez-Leal$^*$, Chao Gao, and Matthew E.~Taylor}
\affiliation{%
 \institution{Borealis AI}
 \city{Edmonton}
 \country{Canada}
}
\email{bilal.kartal, pablo.hernandez, matthew.taylor@borealisai.com}

\begin{abstract}
Safe reinforcement learning has many variants and it is still an open research problem. Here, we focus on how to use action guidance by means of a non-expert demonstrator to avoid catastrophic events in a domain with sparse, delayed, and deceptive rewards: the recently-proposed multi-agent benchmark of Pommerman. This domain is very challenging for reinforcement learning (RL) --- past work has shown that model-free RL algorithms fail to achieve significant learning. In this paper, we shed light into the reasons behind this failure by exemplifying and analyzing the high rate of catastrophic events (i.e., suicides) that happen under random exploration in this domain. While model-free random exploration is typically futile, we propose a new framework where even a non-expert simulated demonstrator, e.g., planning algorithms such as Monte Carlo tree search with small number of rollouts, can be integrated to asynchronous distributed deep reinforcement learning methods. Compared to vanilla deep RL algorithms, our proposed methods both learn faster and converge to better policies on a two-player mini version of the Pommerman game.
\end{abstract}


\maketitle

\blfootnote{$^*$ Equal contribution}

\section{Introduction}

\noindent Deep reinforcement learning (DRL) has enabled better scalability and generalization for challenging high-dimensional domains. DRL has been a very active area of research in recent years~\cite{arulkumaran2017deep,li2017deep,hernandez2018multiagent} with great successes in Atari games~\cite{mnih2015human}, Go~\cite{silver2017mastering} and very recently, multiagent games (e.g., Starcraft and DOTA 2)~\cite{openfive}.

One of the biggest challenges for deep reinforcement learning is sample efficiency~\cite{yu2018towards}. However, once a DRL agent is trained, it can be deployed to act in real-time by only performing an inference through the trained model. On the other hand, planning methods such as Monte Carlo tree search (MCTS)~\cite{browne2012survey} do not have a training phase, but they perform simulation based rollouts assuming access to a simulator to find the best action to take. One major challenge with vanilla MCTS is the scalability to domains with large branching factors and long episodes requiring lots of simulation-based episodes to act, thus, rendering the method inapplicable for applications requiring real-time decision making.

There are several ways to get the best of both DRL and search methods. For example, AlphaGo~\cite{silver2016mastering} and Expert Iteration~\cite{anthony2017thinking} concurrently proposed the idea of combining DRL and MCTS in an imitation learning framework where both components improve each other. These works combine search and neural networks \emph{sequentially} in a loop. First, search is used to generate an expert move dataset, which is used to train a policy network~\cite{guo2014deep}. Second, this network is used to improve expert search quality~\cite{anthony2017thinking}, and this is repeated. However, expert move data collection by vanilla search algorithms can be slow in a sequential framework depending on the simulator efficiency~\cite{guo2014deep}.

In this paper, complementary to the aforementioned existing work, we show that it is also possible to blend search with distributed model-free DRL methods such that search and neural network components can be executed \emph{simultaneously} in an \textit{on-policy} fashion. The main focus of this work is to show how we can use relatively weaker demonstrators (e.g., lightweight MCTS with a small number of rollouts or other search based planners~\cite{lavalle2006planning}) for \emph{safer} model-free RL by coupling the demonstrator and model-free RL interaction through an auxiliary task~\cite{jaderberg2016reinforcement}.

\begin{figure}
\centering
\includegraphics[scale=0.23]{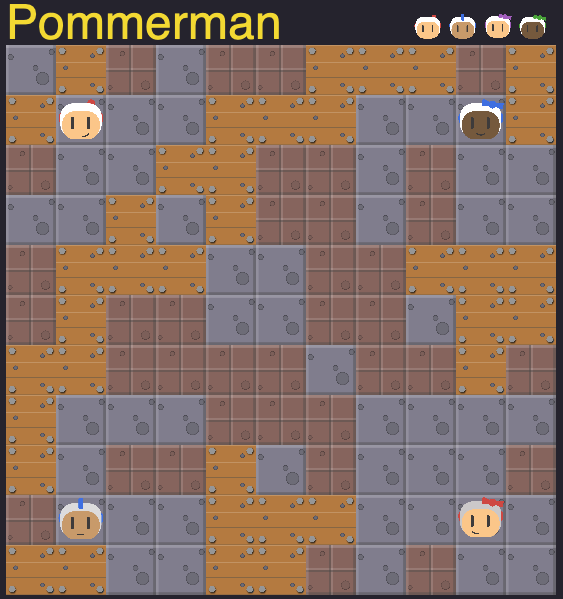}
\caption{A random initial board of Pommerman. Agents have 800 timesteps to blast opponents.
}
\label{fig:pommerman}
\end{figure}

Here we focus on the catastrophic events that arise frequently in a recently proposed benchmark for (multi-agent) reinforcement learning:  Pommerman~\cite{resnick2018pommerman}. This environment is based on the classic console game \emph{Bomberman}. The Pommerman environment involves 4 bomber agents initially placed at the four corners of a board, see Figure~\ref{fig:pommerman}, which take simultaneous actions. %
On the one hand, the only way to make a change in the Pommerman environment (e.g., kill an agent) is by means of bomb placement (and the effects of such an action is only observed when the bomb explodes after 10 time steps). On the other hand, this action could result in the catastrophic event of the agent committing suicide.

In this work we show that suicides happen frequently during learning because of exploration and due to the nature of the environment. Furthermore, the high rate of suicides has a direct effect on the samples needed to learn. We exemplify this in a case for which an \emph{exponential} number of samples are needed to obtain a positive experience. This highlights that performing non-suicidal bomb placement could require complicated, long-term, and accurate planning, which is very hard to learn for model-free reinforcement learning methods.

We consider Asynchronous Advantage Actor-Critic (A3C)~\cite{mnih2016asynchronous} as a baseline algorithm and we propose a new framework based on diversifying some of the workers of A3C with MCTS based planners (serving as non-expert demonstrators) by using the parallelized asynchronous training architecture. This has the effect of providing action guidance, which in turns improves the training efficiency measured by higher rewards.

\section{Related Work}

Our work lies at the intersection of the research areas of safe reinforcement learning, imitation learning, and Monte Carlo tree search based planning. In this section, we will mention some of the existing work in these areas.

\subsection{Safe RL}

Safe Reinforcement Learning tries to ensure reasonable system performance and/or respect safety constraints during the learning and/or deployment processes~\cite{garcia2015comprehensive}. Roughly, there are two ways of doing safe RL: some methods adapt the optimality criterion, while others adapt the exploration mechanism. Since classical approaches to exploration like $\epsilon$-greedy or Boltzmann exploration do not guarantee safety~\cite{leike2017ai,ecoffet2019go}, the research area of \emph{safe exploration} deals with the question -- how can we build agents that respect the safety constraints not only during normal operation, but also during the initial learning period?~\cite{leike2017ai}.

One model-based method proposed to learn the accuracy of the current policy and use another safe policy only in unsafe situations~\cite{ghavamzadeh2016safe}. %
Lipton et al.~\cite{lipton2016combating} proposed learning a risk-assessment model, called \emph{fear} model, which predicts risky events to shape the learning process. Saunders et al.~\cite{saunders2018trial} studied if human intervention could prevent catastrophic events; while the approach was successful in some Atari games, the authors argue it would not scale in more complex environments due to the amount of human labour needed.

\subsection{Imitation Learning} %

Domains where rewards are delayed and sparse are difficult exploration RL problems and are particularly difficult when learning \textit{tabula rasa}. Imitation learning can be used to train agents much faster compared to learning from scratch.

Approaches such as DAgger~\cite{ross2011reduction} or its extended version~\cite{sun2017deeply} formulate imitation learning as a supervised problem where the aim is to match the performance of the demonstrator. However, the performance of agents using these methods is upper-bounded by the demonstrator performance.

Lagoudakis et al.~\cite{lagoudakis2003reinforcement} proposed a classification-based RL method using Monte-Carlo rollouts for each action to construct a training dataset to improve the policy iteratively. Recent works such as Expert Iteration~\cite{anthony2017thinking} extend imitation learning to the RL setting where the demonstrator is also continuously improved during training. There has been a growing body of work on imitation learning where human or simulated demonstrators' data is used to speed up policy learning in RL~\cite{hester2017deep,subramanian2016exploration,cruz2017pre,christiano2017deep,nair2018overcoming}.

Hester et al.~\cite{hester2017deep} used demonstrator data by combining the supervised learning loss with the Q-learning loss within the DQN algorithm to pre-train and showed that their method achieves good results on Atari games by using a few minutes of game-play data. Cruz et al.~\cite{cruz2017pre} employed human demonstrators to pre-train their neural network in a supervised learning fashion to improve feature learning so that the RL method with the pre-trained network can focus more on policy learning, which resulted in reducing training times for Atari games. Kim et al.~\cite{kim2013learning} proposed a learning from demonstration approach where limited demonstrator data is used to impose constraints on the policy iteration phase and they theoretically prove bounds on the Bellman error.

In some domains (e.g., robotics) the tasks can be too difficult or time consuming for humans to provide full demonstrations. Instead, humans can provide \textit{feedback}~\cite{loftin2014strategy,christiano2017deep} on alternative agent trajectories that RL can use to speed up learning. Along this direction, Christiano et al.~\cite{christiano2017deep} proposed a method that constructs a reward function based on data containing human feedback with agent trajectories and showed that a small amount of non-expert human feedback suffices to learn complex agent behaviours.

\subsection{Combining Search, DRL, and Imitation Learning}

AlphaGo~\cite{silver2016mastering} defeated the strongest human Go player in the world on a full-size board. It uses imitation learning by pretraining RL's policy network from human expert games with supervised learning~\cite{lecun2015deep}. Then, its policy and value networks keep improving by self-play games via DRL. Finally, an MCTS search skeleton is employed where a policy network narrows down move selection (i.e., effectively reducing the branching factor) and a value network helps with leaf evaluation (i.e., reducing the number of costly rollouts to estimate state-value of leaf nodes). AlphaGo Zero~\cite{silver2017mastering} dominated AlphaGo even though it started to learn \textit{tabula rasa}. AlphaGo Zero still employed the skeleton of MCTS algorithm, but it employed the value network for leaf node evaluation without any rollouts. Our work can be seen as complementary to these Expert Iteration based methods and differs in multiple aspects: (i) our framework specifically aims to enable on-policy model-free RL methods to explore \emph{safely} in hard-exploration domains where negative rewarding terminal states are ubiquitous, in contrast to off-policy Expert-Iteration based methods which use intensive search to fully learn a policy; (ii) our framework is general in that other demonstrators (human or other simulated sources) can easily be integrated to provide action guidance by using the proposed auxiliary loss refinement; and
(iii) our framework aims to use the demonstrator with a small lookahead (i.e., shallow) search to filter out actions leading to immediate negative terminal states so that model-free RL can imitate those safer actions to learn to safely explore.

\section{Preliminaries}

\subsection{Reinforcement Learning}

 We start with the standard reinforcement learning setting of an agent interacting in an environment over a discrete number of steps. At time $t$ the agent in state $s_t$ takes an action $a_t$ and receives a reward $r_t$. The discounted return is defined as $R_{t:\infty} = \sum_{t=1}^\infty \gamma^t r_t$. The state-value function, $$V^\pi(s)=\mathbb{E}[R_{t:\infty}|s_t=s,\pi]$$ is the expected return from state $s$ following a policy $\pi(a|s)$
and the action-value function is the expected return following policy $\pi$ after taking action $a$ from state $s$:
$$Q^\pi(s,a)=\mathbb{E}[R_{t:\infty}|s_t=s, a_t=a,\pi].$$

The A3C method, as an actor-critic algorithm, has a policy network (actor) and a value network (critic) where the actor is parameterized by $\pi(a|s;\theta)$ and the critic is parameterized by $V(s; \theta_v)$, which are updated as follows:
$$\triangle\theta = \nabla_\theta \log \pi(a_t|s_t; \theta) A(s_t, a_t; \theta_v), $$
$$\triangle\theta_v = A(s_t, a_t; \theta_v) \nabla_{\theta_v} V(s_t)$$
where,
$$A(s_t, a_t; \theta_v) = \sum_k^{n-1} \gamma^kr_{t+k} + \gamma^n V(s_{t+n}) - V (s_t)$$
with $A(s,a)=Q(s,a)-V(s)$ representing the \emph{advantage} function.
The policy and the value function are updated after every $t_{max}$ actions or when a terminal state is reached. It is common to use one softmax output for the policy $\pi(a_t|s_t; \theta)$ head and one linear output for the value function $V (s_t; \theta_v)$ head, with all non-output layers shared.

The loss function for A3C is composed of two terms: policy loss (actor), $\mathcal{L}_{\pi}$, and value loss (critic), $\mathcal{L}_{v}$. An entropy loss for the policy, $H(\pi)$, is also commonly added which helps to improve exploration by discouraging premature convergence to suboptimal deterministic policies~\cite{mnih2016asynchronous}. Thus, the loss function is given by: $$\mathcal{L}_{\text{A3C}} = \lambda_v  \mathcal{L}_{v} + \lambda_{\pi} \mathcal{L}_{\pi} - \lambda_{H} \mathbb{E}_{s \sim \pi} [H(\pi(s, \cdot, \theta)] $$ with $\lambda_{v}=0.5$, $\lambda_{\pi}=1.0$, and $\lambda_{H}=0.01$, being standard weighting terms on the individual loss components.

UNREAL~\cite{jaderberg2016reinforcement} proposed unsupervised \emph{auxiliary tasks} (e.g., reward prediction) to speed up learning, which require no additional feedback from the environment. In contrast to A3C, UNREAL uses an experience replay buffer that is sampled with more priority given to positively rewarded interactions to improve the critic network.
The UNREAL framework optimizes a single combined loss function $\mathcal{L}_{\text{UNREAL}} \approx \mathcal{L}_{A3C} + \lambda_{AT} \mathcal{L}_{AT} $, that combines the A3C loss, $\mathcal{L}_{A3C}$, together with an auxiliary task loss $\mathcal{L}_{AT}$, where $\lambda_{AT}$ is a weight term.

\subsection{Monte Carlo Tree Search}

\begin{figure}
\centering
\includegraphics[width=0.21\linewidth]{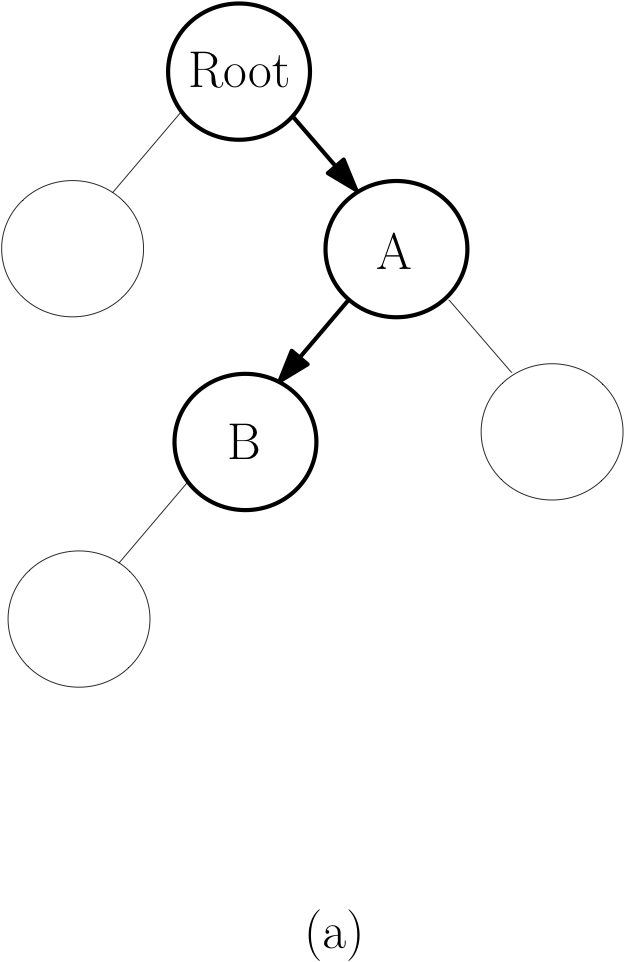}
\label{fig:selection}
\quad
\includegraphics[width=0.21\linewidth]{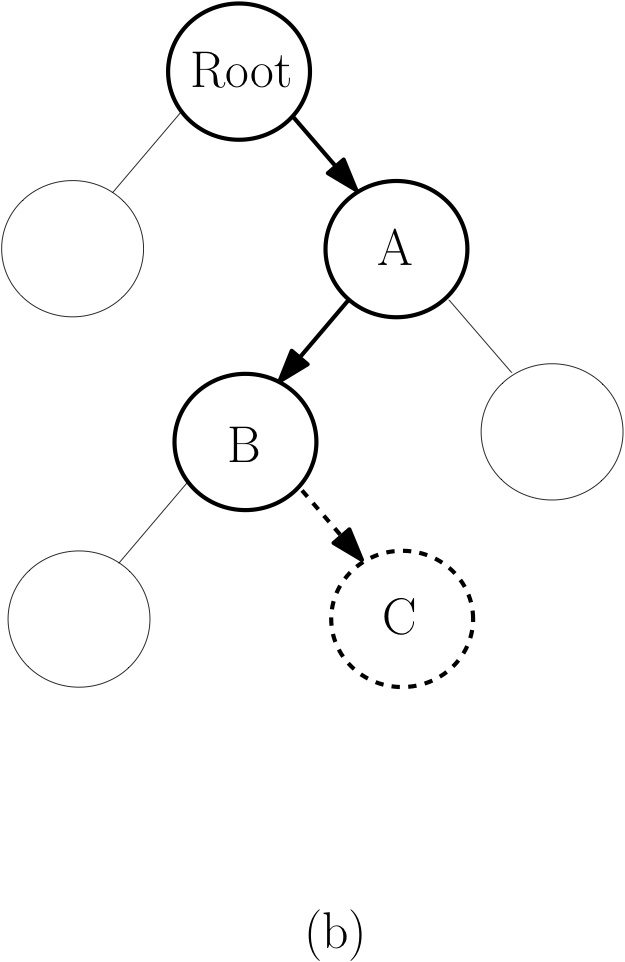}
\label{fig:expansion}
\quad
\includegraphics[width=0.21\linewidth]{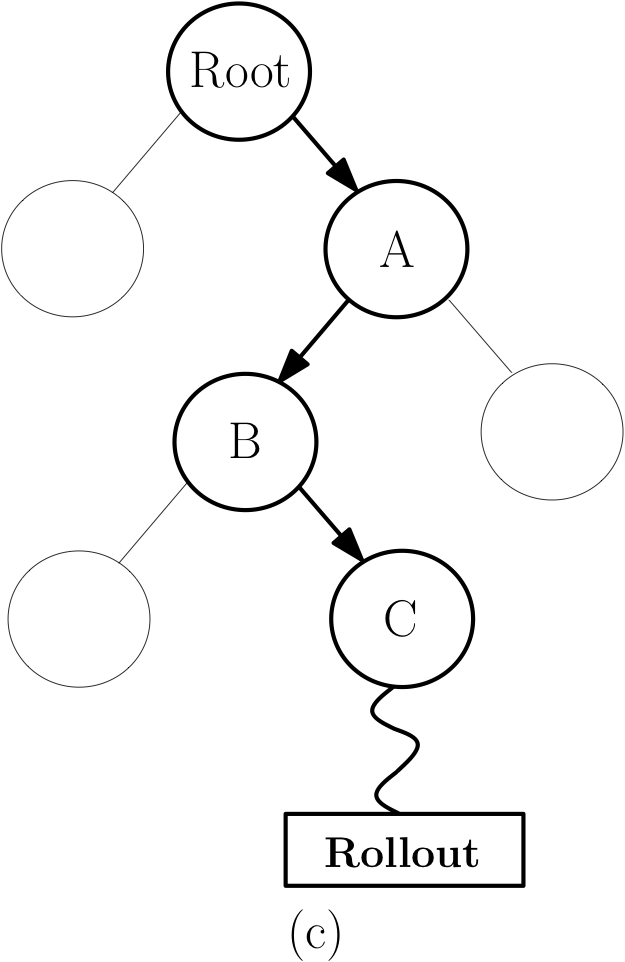}
\label{fig:rollout}
\quad
\includegraphics[width=0.21\linewidth]{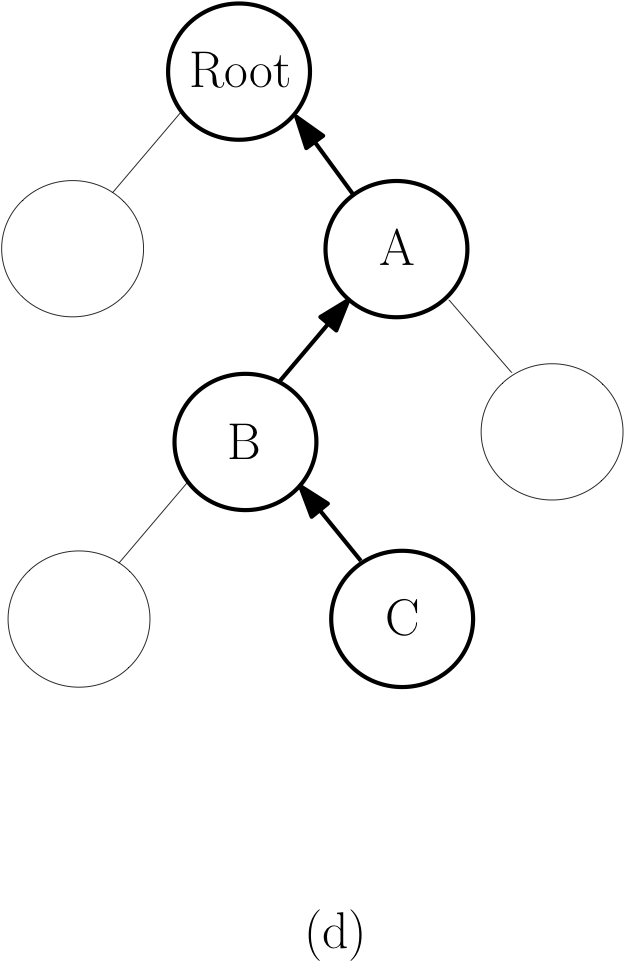}
\label{fig:backprop}
\quad
\caption{ Overview of Monte Carlo Tree Search:
(a) \textit{Selection}: UCB is used recursively until a node with an unexplored action is selected. Assume that nodes A and B are selected.
(b) \textit{Expansion}: Node C is added to the tree.
(c) \textit{Random Rollout}: A sequence of random actions is taken from node C to complete the partial game.
(d) \textit{Back-propagation}: After the rollout terminates, the game is evaluated and the score is back-propagated from node C to the root.}
\label{fig:MCTS}
\end{figure}

Monte Carlo Tree Search is a best-first search algorithm that gained traction after its breakthrough performance in Go \cite{coulom2006efficient}. Other than for game playing agents~\cite{sturtevant2015monte,bravi2018shallow} and playtesting agents~\cite{zook2015monte,holmgaard2018automated,borovikov2019winning}, MCTS has been employed for a variety of domains such as robotics~\cite{zhang2017active}, continuous control tasks for animation~\cite{rajamaki2018continuous}, and procedural puzzle generation~\cite{kartal2016data}.  One recent paper~\cite{vodopivec2017monte} provided an excellent unification of MCTS and RL.

In MCTS a search tree is generated where each node in the tree represents a complete state of the domain and each link represents one possible action from the set of valid actions in the current state, leading to a child node representing the resulting state after applying that action. The root of the tree is the initial state (for example, the initial configuration of the Pommerman board including the agent location).
MCTS proceeds in four phases of: selection, expansion, rollout, and back-propagation (see Figure~\ref{fig:MCTS}).
The MCTS algorithm proceeds by repeatedly adding one node at a time to the current tree. Given that actions from the root to the expanded node is unlikely to terminate an episode, e.g., a Pommerman game can take up to 800 timesteps, MCTS uses random actions, a.k.a. \textit{rollouts}, to estimate state-action values. After rollout phase, the total collected rewards during the episode is back-propagated through all existing nodes in the tree updating their empirical state-action value estimates.

\paragraph{Exploration vs.~Exploitation Dilemma}

Choosing which child node to expand (i.e., choosing which action to take) becomes an exploration/exploitation problem. We want to primarily choose actions that have good scores, but we also need to explore other possible actions in case the observed empirical average scores do not represent the true reward mean of that action. This exploration/exploitation dilemma has been well studied in other areas.
Upper Confidence Bounds (UCB)~\cite{auer2002finite} is a selection algorithm that seeks to balance the exploration/exploitation dilemma. Using UCB with MCTS is also referred to as Upper Confidence bounds applied to Trees (UCT). Applied to our framework, each parent node $s$ chooses its child $s'$ with the largest $UCB(s,a)$ value according to Eqn.~\ref{eqn:ucb}.

\begin{equation}
UCB(s,a) = Q(s,a) + C \sqrt{\frac{\ln n(s)}{n(s')}}
\label{eqn:ucb}
\end{equation}

\noindent Here, $n(s)$ denotes number of visits to the node $s$ and $n(s')$ denotes the number of visits to $s'$, i.e., the resulting child node when taking action $a$ from node $s$. The value of $C$ determines the rate of exploration, where smaller $C$ implies less exploration. Kocsis and Szepesvari~\cite{kocsis2006bandit} showed that $C = \sqrt{2}$ is necessary for asymptotic convergence, however it can be tuned depending on the domain. In this work, we employed the default value of $\sqrt{2}$ for the exploration constant.

\section{Safe RL with MCTS Action Guidance}

\begin{figure*}
\centering
\includegraphics[width=\linewidth]{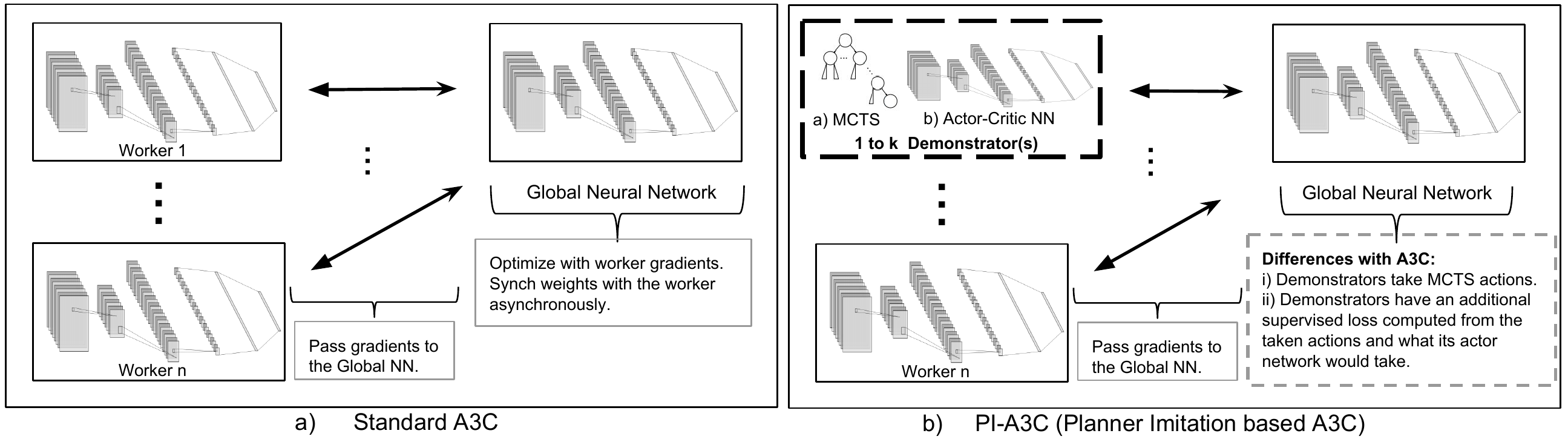}
\caption{\textbf{a)} In the A3C framework, each worker independently interacts with the environment and computes gradients. Then, each worker \emph{asynchronously} passes the gradients to the global neural network which updates parameters and synchronizes with the respective worker. \textbf{b)} In our proposed framework, Planner Imitation based A3C (PI-A3C), $k\ge 1$ CPU workers are assigned as MCTS based demonstrators taking MCTS actions, while keeping track of what action its actor network would take. The demonstrator workers have an additional auxiliary supervised loss. PI-A3C enables the network to simultaneously optimize the policy and learn to imitate the MCTS.}
\label{fig:pi_a3c}
\end{figure*}

In this section, firstly, we present our framework PI-A3C (Planner Imitation - A3C) that extends A3C with a lightweight search based demonstrator through an auxiliary task based loss refinement as depicted in Figure~\ref{fig:pi_a3c}. Secondly, we present the Pommerman domain in detail to analyze and exemplify the exploration complexity of that domain for pure model-free RL methods.

\subsection{Approach Overview}

We propose a framework that can use planners, or other sources of demonstrators, along with asynchronous DRL methods to accelerate learning.
Even though our framework can be generalized to a variety of planners and distributed DRL methods, we showcase our contribution using MCTS and A3C. In particular, our use of MCTS follows the approach of Kocsis and Szepesvari~\cite{kocsis2006bandit} employing UCB to balance exploration versus exploitation during planning.  During rollouts, we simulate all agents as random agents as in default unbiased MCTS and we perform limited-depth rollouts to reduce action-selection time.

The motivation for combining MCTS and asynchronous DRL methods stems from the need to improve training time efficiency even if the planner or environment simulator is very slow. In this work, we assume the demonstrator and actor-critic networks are decoupled, i.e., a vanilla UCT planner is used as a black-box that takes an observation and returns an action resembling UCTtoClassification method~\cite{guo2014deep}, which uses MCTS to generate an expert move dataset and trains a NN in a supervised fashion to imitate MCTS.
However, they used a high-number of rollouts (10K) per action selection to construct an expert dataset. In contrast, we show how vanilla MCTS with a small number of rollouts\footnote{In Pommerman 10K rollouts would take hours due to very slow simulator and long horizon~\cite{matiisen2018pommerman}.} ($\approx$ 100) can still be employed in an \textit{on-policy} fashion to improve training efficiency for actor-critic RL in challenging domains with abundant easily reachable terminal states with negative rewards.  %

Within A3C's asynchronous distributed architecture, all the CPU workers perform agent-environment interaction with their neural network policy networks, see Figure~\ref{fig:pi_a3c}(a). In our new framework, PI-A3C (Planner Imitation with A3C), we assign $k\ge 1$ CPU
workers (we present experimental results with different $k$ values in Section~\ref{sec:results}) to perform MCTS based planning for agent-environment interaction based on the agent's observations, while also keeping track of what its neural network would perform for those observations, see Figure~\ref{fig:pi_a3c}(b). In this fashion, we both learn to imitate the MCTS planner and to optimize the policy. The main motivation for PI-A3C framework is to increase the number of agent-environment interactions with positive rewards for hard-exploration RL problems to improve training efficiency.

Note that the planner-based worker still has its own neural network with actor and policy heads, but action selection is performed by the planner while its policy head is used for loss computation. In particular, the MCTS planner based worker augments its loss function with the auxiliary task of \emph{Planner Imitation}\footnote{Both Guo et al.~\cite{guo2014deep} and Anthony et al.~\cite{anthony2017thinking} used MCTS moves as a learning target, referred to as \textit{Chosen Action Target}. Our \textit{Planner Imitation} loss is similar except in this work, we employed cross-entropy loss in contrast to a KL divergence based one.}. The auxiliary loss is defined as ${\mathcal{L}_{PI}= -\frac{1}{N} \sum_i^N a^i_{o} \log (\hat{a}^i_{o}) }$,  which is the supervised cross entropy loss between  $a^i_{o}$ and $\hat{a}^i_{o}$, representing the one-hot encoded action the planner used and the action the actor (with policy head) would take for the same observation respectively during an episode of length $N$.
 The demonstrator worker's loss after the addition of \emph{Planner Imitation} is defined by $$\mathcal{L}_{\text{PI-A3C}}= \mathcal{L}_{A3C} + \lambda_{PI} \mathcal{L}_{PI} $$
\noindent where $\lambda_{PI}=1$ is a weight term (which was not tuned).
In PI-A3C the rest of the workers (not demonstrators) are left unchanged, still using the policy head for action selection with the unchanged loss function. By formulating the \textit{Planner Imitation} loss as an auxiliary loss, the objective of the resulting framework becomes a multi-task learning problem~\cite{caruana1997multitask} where the agent aims to learn to both maximize the reward and imitate the planner.

\subsection{Pommerman}

In Pommerman, each agent can execute one of 6 actions at every timestep: move in any of four directions, stay put, or place a bomb. Each cell on the board can be a passage, a rigid wall, or wood. The maps are generated randomly, albeit there is always a guaranteed path\footnote{Although this path can be initially blocked by wood, thus, needing clearance by bombs.} between any two agents.
Whenever an agent places a bomb it explodes after 10 timesteps, producing flames that have a lifetime of 2 timesteps. Flames destroy wood and kill any agents within their blast radius. When wood is destroyed either a passage or a power-up is revealed. Power-ups can be of three types: increase the blast radius of bombs, increase the number of bombs the agent can place, or give the ability to kick bombs. A single game is finished when an agent dies or when reaching 800 timesteps.

Pommerman is a challenging benchmark for multi-agent learning and model-free reinforcement learning, due to the following characteristics:

\textbf{Multiagent component:} the agent needs to best respond to any type of opponent, but agents' behaviours also change based on the collected power-ups, i.e., extra ammo, bomb blast radius, and bomb kick ability.

\textbf{Delayed action effects}: the only way to make a change to the environment (e.g., kill an agent) is by means of bomb placement, but the effect of such an action is only observed when the bombs explodes after 10 time steps.

\textbf{Sparse and deceptive rewards}: the former refers to the fact that the only non-zero reward is obtained at the end of an episode. The latter refers to the fact that quite often a winning reward is due to the opponents' involuntary \emph{suicide}, which makes reinforcing an agent's action based on such a reward \emph{deceptive}.

For these reasons, we consider this game challenging for many standard RL algorithms and a local optimum is commonly learned, i.e., not placing bombs~\cite{resnick2018pommerman}.

Some other recent works also used Pommerman as a test-bed, for example, Zhou et al.~\cite{zhou2018hybrid} proposed a hybrid method combining rule-based heuristics with depth-limited search. Resnick et al.~\cite{resnick2018backplay} proposed a framework that uses a single demonstration to generate a training curriculum for sparse reward RL problems (assuming episodes can be started from arbitrary states).  Lastly, relevance graphs, which represent the relationship between agents and environment
objects~\cite{malysheva2018deep}, are another approach to deal with the complex Pommerman game.

\subsection{Catastrophic events in Pommerman: Suicides}

Before presenting the experimental results of our approach we motivate the need for safe exploration by providing two examples and a short analysis of the catastrophic events that occur in Pommerman.

It has been noted that in Pommerman the action of bomb placing is highly correlated to losing~\cite{resnick2018pommerman}. This is presumably a major impediment for achieving good results using model-free reinforcement learning. Here, we provide an analysis of the suicide problem that delays or even prevents to learn the \emph{bombing skill} when an agent follows a random exploration policy.

In Pommerman an agent can only be killed when it intersects with an exploding bomb's flames, then we say a \emph{suicide} event happens if the death of the agent is caused by a previously placed own bomb. For the ease of exposition we consider the following simplified scenario: (i) the agent has \texttt{ammo=1} and has just placed a bomb (i.e., the agent now sits on the bomb). (ii) For the next time steps until the bomb explodes the agent has $5$ actions available at every time step (move in 4 directions or do nothing); and (iii) all other items on the board are static.

\begin{figure}
    \centering
        \centering
        \includegraphics[scale=1.00]{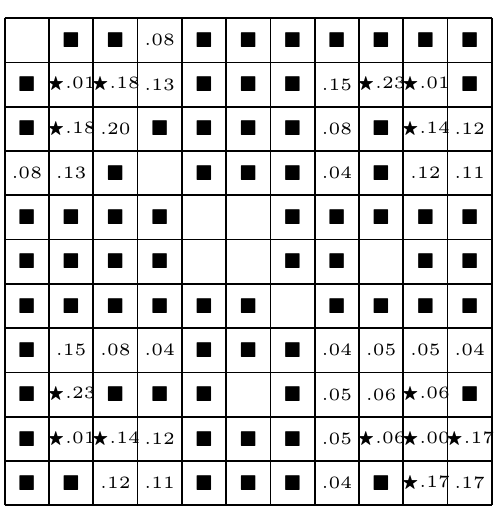}
        \\
    \caption{Probabilities of being at each cell for $t=9$ (after a bomb explodes) for each agent. $\star$ indicates the cell is covered by flames and $\blacksquare$ represents an obstacle, which can be wood or a rigid cell. For each agent, the probabilities of suicide are $\approx 0.4$. Survival probability after placing $b$ bombs is computed by $(1-0.4)^b$, showing that it quickly approaches $0$ (similar the probability of getting $b$ heads in $b$ consecutive coin flips), showing high exploration difficulty. }
    \label{fig:pommerman_prob9}
\end{figure}

\paragraph{Example 1.}
We take as example the board configuration depicted in Figure~\ref{fig:pommerman} and show the probability of ending in a certain position for $t=9$ (which is the time step that the bomb will explode) for the 4 agents, see Figure~\ref{fig:pommerman_prob9}. This is a typical starting board in Pommerman, where every agent stays in its corner; they are disconnected with each other by randomly generated wood and walls. From this figure we can see that for each of the four agents, even when their configuration is different, their probabilities of ending up with suicide are $\approx 40\%$ ($0.39, 0.38, 0.46, 0.38$, respectively, in counter clockwise order starting from upper-left corner) --- these are calculated by summing up the positions where there is a flame, represented by $\star$.

Indeed, the problem of suicide stems from acting randomly without considering constraints of the environment. In extreme cases, in each time step, the agent may have only one survival action, which means the safe path is always unique as $t$ increases while the total number grows exponentially. We illustrate this by the following corridor example.

\paragraph{Example 2.}
Figure~\ref{fig:corridor} shows a worst-case example: the agent is in a corridor formed by wood at two sides and places a bomb. If using random exploration the chance of ending up in suicide is extremely high since among the $5^9$ ``paths,'' i.e., action trajectories ---  \emph{only one} of them is safe. In order to survive, it must precisely follow the right action at each time step.

Note that in cases like the corridor scenario even if the agent is modestly model-aware, i.e., it may only look one step-ahead, the number of possible action trajectories is still exponential, while the survival path remains unique.
This also implies that for such sub-problems in Pommerman, acquiring one positive behaviour example requires \emph{exponential} number of samples.

\begin{figure}[t] \small
    \centering
    \begin{tikzpicture}[scale=0.6]
    \draw (0,0) grid (10, 1);
    \draw (9,1) rectangle (10,2);
    \draw (0.5,0.5) node[shape=circle, opacity=0.4, fill=black, text opacity=1, draw, inner sep=0pt] {${~~9~~}$};
    \foreach \x in {1,...,9}{
        \pgfmathtruncatemacro{\label}{10-\x}
        \draw (\x+0.5,0.5) node[shape=circle, color=blue] {\label};
    };
    \draw (9.5,1.5) node[shape=circle, color=blue] {0};
    \end{tikzpicture}
    \caption{The corridor scenario: the agent places a bomb with \texttt{strength=10} on the leftmost cell. For each passage cell, the marked value means the minimum number of steps it is required to safely evade from impact of the bomb. After placing the bomb, in the next step the bomb has life of $9$, thus in the remaining $9$ steps, the agent must always take the unique correct action to survive. }
    \label{fig:corridor}
\end{figure}

\section{Experiments and Results}

In this section, we present the experimental setup and results against different opponents in a simplified version of the Pommerman game. We run ablation experiments on the contribution of \emph{Planner Imitation} to compare against the standard A3C method for: (i) single demonstrator with different expertise levels, (ii) different number of demonstrators with the same expertise level, and (iii) using rollout biasing within MCTS in contrast to uniform random rollout policy.

All the training curves are obtained from 3 runs with different random seeds. From these 3 separate runs, we compute average learning performance (depicted with bold color curves) and standard deviations (depicted with shaded color curves). At every training episode, the board is randomized ,i.e., the locations of wood and rigid cells are shuffled.

\subsection{Setup}

\begin{figure}
\centering
\includegraphics[scale=0.25]{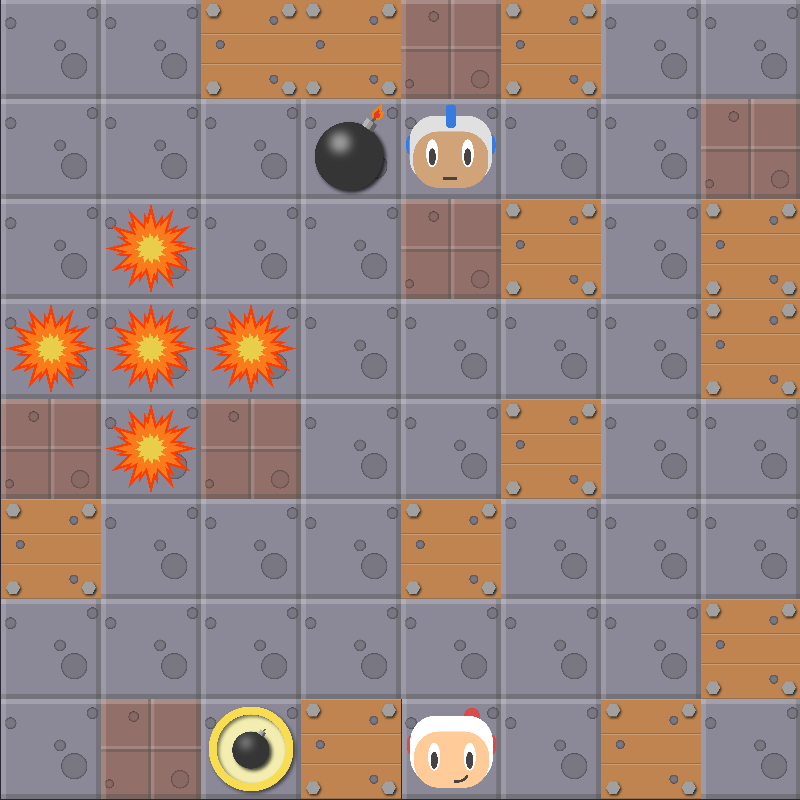}
\caption{An example of the $8 \times 8$ Mini-Pommerman board, randomly generated by the simulator. Agents' initial positions are randomized among four corners at each episode.}
\label{fig:pom8x8}
\end{figure}

Because of all the complexities mentioned in this domain we simplified it by considering a game with only two agents and a reduced board size of $8 \times 8$, see Figure~\ref{fig:pom8x8}. Note that we still randomize \emph{the location of} walls, wood, power-ups, and the initial position of the agents for every episode. We considered two types of opponents in our experiments:

\begin{itemize}
    \item \emph{Static} opponents: the opponent waits in the initial position and always executes the `stay put' action. This opponent provides the easiest configuration (excluding suicidal opponents). It is a baseline opponent to show how challenging the game is for model-free RL.

    \item \emph{Rule-based} opponents: this is the benchmark agent within the simulator. It collects power-ups and places bombs when it is near an opponent. It is skilled in avoiding blasts from bombs. It uses Dijkstra's algorithm on each time-step, resulting in longer training times.

\end{itemize}

Details regarding neural network architecture and implementation are provided in the appendix.

\subsection{Results}
\label{sec:results}

\begin{figure*}[t]
    \subfloat[Learning against a Static Agent]{{\includegraphics[scale=0.465]{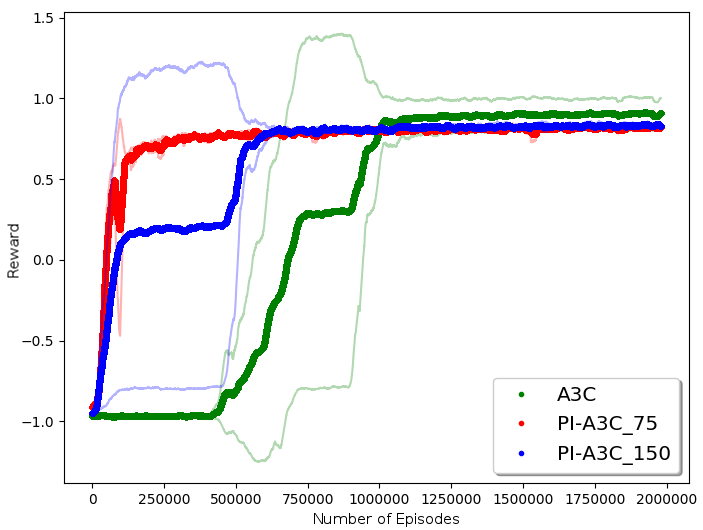} }}%
    \subfloat[Learning against a Rule-based Agent]{{\includegraphics[scale=0.42]{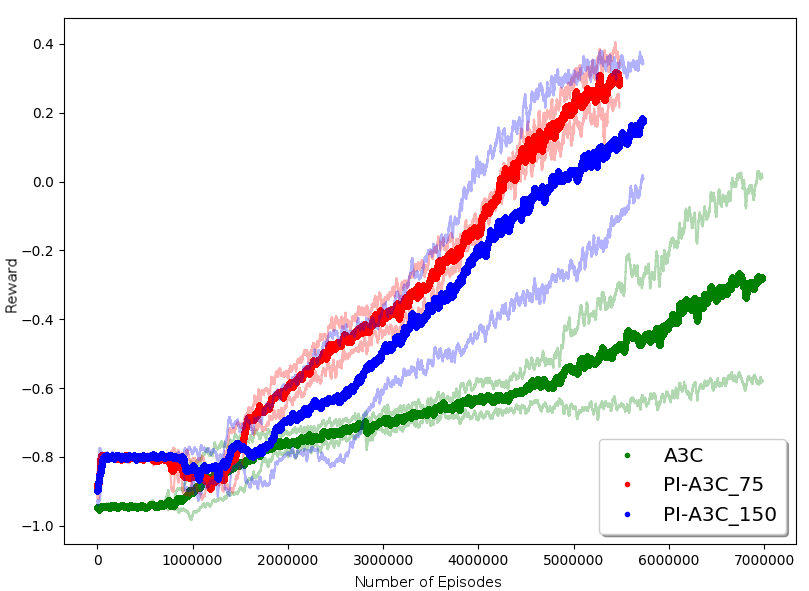} }}%
    \caption{Both figures are obtained from 3 training runs showing the learning performance with mean (bold lines) and standard deviation (shaded lines). 24 CPU workers are used for all experiments, and all PI-A3C variants use only 1 Demonstrator worker. a) Against Static agent, all variants have been trained for 12 hours. The ${\text{PI-A3C}}$ framework using MCTS demonstrator with 75 and 150 rollouts learns significantly faster compared to the standard A3C. b) Against Rule-based opponent, all variants have been trained for 3 days. Against this more skilled opponent, PI-A3C provides significant speed up in learning performance, and finds better best response policies. For both (a) and (b), increasing the expertise level of MCTS through doubling the number of rollouts (from 75 to 150) does not yield improvement, and even can hurt performance. Our hypothesis is that slower planning decreases the number of demonstrator actions too much for the model-free RL workers to learn to imitate for safe exploration.}%
    \label{fig:ablate_planner}%
\end{figure*}

We conducted two sets of experiments learning against \emph{Static} and \emph{Rule-based} opponents. We compare our proposed methods with respect to standard A3C with learning curves in terms of converged policies and time-efficiency. All approaches were trained using 24 CPU cores. For a fair comparison all training curves for PI-A3C variants are obtained by using only the neural network policy network based workers, excluding rewards obtained by the demonstrator worker to accurately observe the performance of model-free RL.

\paragraph{On action guidance, quantity versus quality}

Within our framework, we can vary the expertise level of MCTS by simply changing the number of rollouts games per move for the demonstrator worker. We experimented with 75 and 150 rollouts. Given finite training time, higher rollouts imply deeper search and better moves, however, it also implies that number of guided actions by the demonstrators will be fewer in quantity, reducing the number of asynchronous updates to the global neural network. As Figure~\ref{fig:ablate_planner} shows against both opponents the relatively weaker demonstrator (75 rollouts) enabled faster learning than the one with 150 rollouts. We hypothesize that the faster demonstrator (MCTS with only 75 rollouts) makes more updates to the global neural network, warming up other purely model-free workers for \emph{safer} exploration much earlier in contrast to the slower demonstrator. This is reasonable as the model-free RL workers constitute all but one of the CPU workers in these experiments, therefore the earlier the model-free workers can start safer exploration, the better the learning progress is likely to be.\footnote{Even though we used MCTS with fixed number of rollouts, this could be set dynamically, for example, by exploiting the reward sparsity or variance specific to the problem domain, e.g., using higher number of rollouts when close to bombs or opponents.}

\paragraph{On the trade-off of using multiple demonstrators}

\begin{figure}
\includegraphics[scale=0.32]{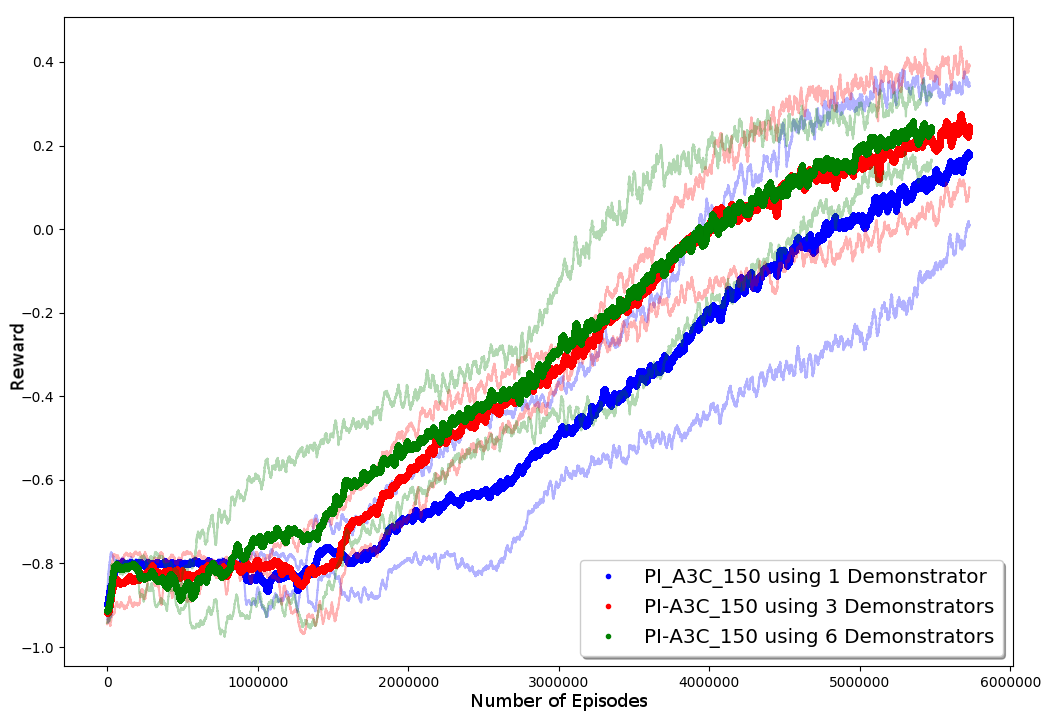}
\caption{ Learning against Rule-based opponent. Employing different number $(n=1,3,6)$ of Demonstrators within the asynchronous distributed framework. Increasing from 1 to 3 the number of demonstrators  also improved the results, however, there is almost no variation from 3 to 6 demonstrators. They all outperform standard A3C, see Figure~\ref{fig:ablate_planner}(b).}
\label{fig:simple_different_demonstrators}
\end{figure}

Our proposed method is built on top of an asynchronous distributed framework that uses several CPU workers: $k\ge 1$ act as a demonstrator and the rest of them explore in model-free fashion. We conducted one  experiment to better understand how increasing the number of demonstrators, each of which provides additional \emph{Planner Imitation} losses asynchronously, affects the learning performance. The trade-off is that more demonstrators imply fewer model-free workers to optimize the main task, but also a higher number of actions to imitate. We present the results in Figure~\ref{fig:simple_different_demonstrators} where we experimented 3 and 6 demonstrators, with identical resources and with 150 rollouts each. Results show that increasing to 3 improves the performance while extending to 6 demonstrators does not provide any marginal improvement. We can also observe that 3 demonstrator version using 150 rollouts presented in Figure~\ref{fig:simple_different_demonstrators} has a relatively similar performance with the 1 demonstrator version using 75 rollouts depicted in Figure~\ref{fig:ablate_planner}(b), which is aligned with our hypothesis that providing more demonstrator guidance early during learning is more valuable than fewer higher quality demonstrations.

\paragraph{Demonstrator biasing with policy network}

\begin{figure}
\centering
\includegraphics[scale=0.48]{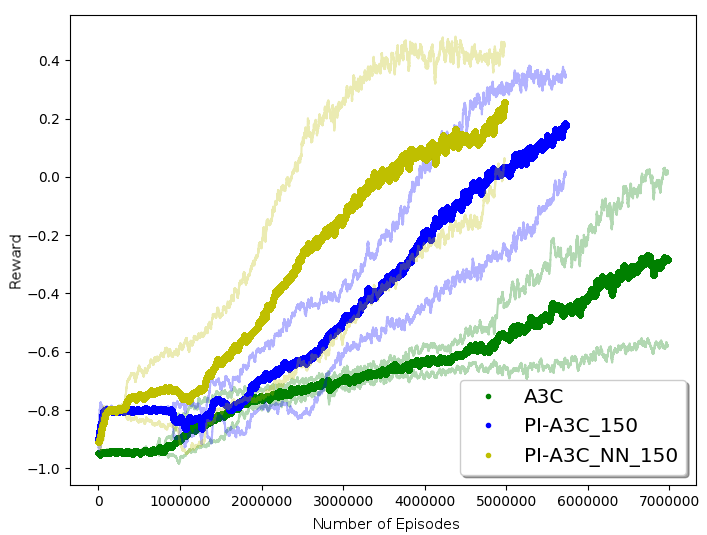}
\caption{Learning against Rule-based opponent. Employing policy network during MCTS rollout phase within the demonstrator provides improvement in learning speed, but it has higher variance compared to employing the standard uniform random rollouts.}
\label{fig:simple_rollout_biasing}
\end{figure}

Uniform random rollouts employed by vanilla MCTS to estimate state-action values provide an unbiased estimate, however due to the high variance, it requires many rollouts. One way to improve search efficiency has been through different biasing strategies rather than using uniform rollout policy, such as prioritizing actions globally based on their evaluation scores~\cite{kartal2014user}, using heuristically computed move urgency values~\cite{bouzy2005associating}, or concurrently learning a rollout policy RL~\cite{ilhan2017monte}. In a similar vein with these methods, we let the MCTS based demonstrator to use the policy network during the rollout phase. We name this refined version as PI-A3C-NN (Planner Imitation - A3C with Neural Network). Our results suggest that employing a biased rollout policy provides improvement in the average learning performance, however it has higher variance as depicted in Figure~\ref{fig:simple_rollout_biasing}.

\section{Discussion}

In Pommerman, the main challenge for model-free RL is the high probability of suicide while exploring (yielding a negative reward), which occurs due to the delayed bomb explosion. However, the agent cannot succeed without learning how to stay safe after bomb placement. The methods and ideas proposed in this paper address this hard-exploration challenge. The main idea of our proposed framework is to use MCTS as a shallow demonstrator (small number of rollouts). This yielded fewer training episodes with catastrophic suicides as imitating MCTS provided improvement for a safer exploration. Concurrent work has proposed the use of pessimistic scenarios to constraint real-time tree search~\cite{osogami2019real}. Other recent work~\cite{lee2019wisemove} employed MCTS as a high-level planner, which is fed a set of low-level offline learned DRL policies and refines them for safer execution within a simulated autonomous driving domain.

There are several directions to extend our work. One direction is to investigate how to formulate the problem so that \emph{ad-hoc} invocation of an MCTS based simulator can be employed. Currently, MCTS provides a safe action continuously, but action guidance can be employed only when the model-free RL agent indeed needs one, e.g., in Pommerman whenever a bomb is about to go off or near enemies where enemy engagement requires strategic actions.

All of our work presented in the paper is on-policy for better comparison to the standard A3C method --- we maintain no experience replay buffer. This means that MCTS actions are used only once to update neural network and thrown away.  In contrast, UNREAL uses a buffer and gives higher priority to samples with positive rewards. We could take a similar approach to save demonstrator's experiences to a buffer and sample based on the rewards.

\section{Conclusions}

Safe reinforcement learning has many variants and it is still an open research problem. In this work, we present a framework that uses a non-expert simulated demonstrator within a distributed asynchronous deep RL method to succeed in hard-exploration domains. Our experiments use the recently proposed Pommerman domain, a very challenging benchmark for pure model-free RL methods as the rewards are sparse, delayed, and deceptive. We provide examples of these issues showing that RL agents fail mainly because the main skill to be learned is highly correlated with negative rewards due to the high-probability of catastrophic events. In our framework, model-free workers learn to safely explore and acquire this skill by imitating a shallow search based non-expert demonstrator. We performed different experiments varying the quality and the number of demonstrators. The results shows that our proposed method shows significant improvement in learning efficiency across different opponents.

\vspace{1cm}
\section*{Appendix}

\textbf{Neural Network Architecture:} For all methods described in the paper, we use a deep neural network with 4 convolutional layers, each of which has 32 filters and $3 \times 3$ kernels, with stride and padding of 1, followed with 1 dense layer with 128 hidden units, followed with 2-heads for actor and critic (where the actor output corresponds to probabilities of 6 actions, and the critic output corresponds to state-value estimate). Neural network architectures were not tuned.

\textbf{NN State Representation:} Similar to ~\cite{resnick2018pommerman}, we maintain 28 feature maps that are constructed from the agent observation. These channels maintain location of walls, wood, power-ups, agents, bombs, and flames. Agents have different properties such as bomb kick, bomb blast radius, and number of bombs. We maintain 3 feature maps for these abilities per agent, in total 12 is used to support up to 4 agents. We also maintain a feature map for the remaining lifetime of flames. All the feature channels can be readily extracted from agent observation except the opponents' properties and the flames' remaining lifetime, which can be tracked efficiently by comparing sequential observations for fully-observable scenarios.

\textbf{Hyperparameter Tuning:} We did not perform a through hyperparameter tuning due to long training times. We used a $\gamma=0.999$ for discount factor. For A3C, the default weight parameters are employed, i.e., $1$ for actor loss, $0.5$ for value loss, and $0.01$ for entropy loss. For the \emph{Planner Imitation} task, $\lambda_{PI}=1$ is used for the MCTS worker, and $\lambda_{PI}=0$ for the rest of workers. We employed the Adam optimizer with a learning rate of $0.0001$. We found that for the Adam optimizer, $\epsilon = 1\times10^{-5}$ provides a more stable learning curve (less catastrophic forgetting) than its default value of $1\times10^{-8}$. We used a weight decay of $1\times10^{-5}$ within the Adam optimizer for L2 regularization.

\bibliographystyle{ACM-Reference-Format}  %

\bibliography{ref}  %

\end{document}